%% file: amia.tex
\documentclass{amia}
\usepackage{lipsum} %Remove if not needed
\usepackage{xcolor}
\usepackage{todonotes}
\usepackage{multirow}

\begin{document}

\title{PLACID: Privacy-preserving Large language models for Acronym Clinical Inference and Disambiguation}

% \author{Firstname A. Lastname, MD, MPH$^1$, Firstname B. Lastname, MD, PhD$^2$ }
\author{Manjushree B. Aithal, Ph.D.$^1$, Alexander Kotz$^1$, Suraju A Sadeeq$^1$, James Mitchell, Ph.D.$^1$ }

\institutes{
    $^1$ Department of Biomedical Informatics, University of Colorado Anschutz, Aurora, CO
}

\maketitle

\section*{Abstract}

\textit{Large Language Models (LLMs) offer transformative solutions across many domains, but healthcare integration is hindered by strict data privacy constraints. Clinical narratives are dense with ambiguous acronyms, misinterpretation these abbreviations can precipitate severe outcomes like life-threatening medication errors. While cloud-dependent LLMs excel at Acronym Disambiguation, transmitting Protected Health Information to external servers violates privacy frameworks. To bridge this gap, this study pioneers the evaluation of small-parameter models deployed entirely on-device to ensure privacy preservation.
%fully localized, small parameter models deployed entirely on-device to guarantee data security. 
We introduce a privacy-preserving cascaded pipeline leveraging general-purpose local models to detect clinical acronyms, routing them to domain-specific biomedical models for context-relevant expansions. Results reveal that while general instruction-following models achieve high detection accuracy (\textbf{$\sim$0.988}), their expansion capabilities plummet (\textbf{$\sim$0.655}). Our cascaded approach utilizes domain-specific medical models to increase expansion accuracy to (\textbf{$\sim$0.81}). 
%By providing self-assessed confidence scores, 
This novel work demonstrates that privacy-preserving, on-device (2B-10B) models deliver high-fidelity clinical acronym disambiguation support.
}

\textbf{Key words}: Natural Language Processing, Privacy and Security, Large Language Models (LLMs)
%Clinical Natural Language Processing (NLP), Acronym Disambiguation, Small Language Models (SLMs), Privacy-Preserving AI

%Abstract text goes here, justified and in italics.  The abstract would normally be one paragraph long. See Table~\ref{tab:submission}. for appropriate abstract length by submission type.

\input{introduction}
\input{related_work}
\input{experiment_setup}
\input{observations}

\input{disucssion}

\input{conclusion}
\makeatletter
\renewcommand{\@biblabel}[1]{\hfill #1.}
\makeatother

% unstr is used to keep citation order
\bibliographystyle{vancouver}
\bibliography{amia}

\end{document}

%% file: introduction.tex
\section*{Introduction}

% Clinical text is characterized by rapid documentation, resulting in a high density of abbreviations, acronyms, and specialized shorthand. While efficient for the author, this abbreviated language is inherently obscure. As Kuhn~\cite{kuhn} warned about "shorter isn’t sweeter," and the misinterpretation of clinical acronyms remains a persistent source of preventable medical errors and patient harm (Amosa et al.,~\cite{amosa2023clinical}). Berger~\cite{} through their findings further underscored that effective communication is essential in intensive-care units, yet the literature shows that abbreviations and acronyms are frequently ambiguous and can jeopardize patient safety. The authors set out to catalog the most ambiguous shortcuts used in critical-care settings and evaluated their potential impact on clinical errors and communication quality. Misinterpretation can precipitate medical errors, delayed treatments

%The large volume of ambiguous acronyms and abbreviations in critical-care documentation represents a clear patient-safety hazard. Misinterpretation can precipitate medication errors, delayed treatment, and inappropriate interventions potentially jeopardizing patient safety~\cite{berger2026shortcuts}. 
The large volume of ambiguous acronyms and abbreviations in critical-care documentation represents a clear patient-safety hazard~\cite{berger2026shortcuts}. Misinterpretation can precipitate medication errors, delayed treatment, and inappropriate interventions potentially jeopardizing patient safety. As an example from Berger's~\cite{berger2026shortcuts} study, the authors describe an incident where the abbreviation ``CA" (intended as ``cancer") was incorrectly interpreted by unfamiliar clinician as ``cardiac arrest". This triggered inappropriate resuscitation efforts and harmful catecholamine exposure while concurrently delaying essential chemotherapeutic infusions. 
%Prior research highlighted, the misinterpretation of medical shorthand poses a substantial safety hazard. For instance, a retrospective analysis encompassing over 643,000 medication errors across 682 healthcare facilities revealed that 5\% of these adverse events stemmed directly from abbreviation use~\cite{brunetti2007impact}. 
While administrative strategies such as standardized lists and regular clinician education, offer a baseline for risk reduction, however, the rapid pace and high cognitive load of clinical environments make strict manual compliance difficult to sustain. Consequently, there is an urgent need for automated, computational approaches capable of seamlessly disambiguating these terms in real-time, effectively mitigating risk without adding to the provider's documentation burden. Acronym Disambiguation, mapping an abbreviation to its correct expansion based on surrounding context, has become a foundational requirement for downstream healthcare informatics, directly impacting the efficacy of everything from automated medical coding to real-time mortality prediction (Nakayama et al.,~\cite{nakayama2019making}). 

% The computational approach to acronym disambiguation has undergone a profound paradigm shift over the past two decades. Early solutions relied on rigid, knowledge-based pipelines such as MetaMap~\cite{aronson2001effective} or heavily engineered statistical classifiers. Today, advanced deep learning 

While the necessity of acronym disambiguation is well established, the computational approaches to solve it have evolved dramatically. Traditional methodologies often relied on rigid, rule-based dictionaries or early statistical classifiers~\cite{skreta2021automatically, kashyap2020classe, link2020} that struggled to generalize across diverse clinical settings. 
%In recent years, the introduction of Large Language Models (LLMs) has redefined the state-of-the-art. 
In recent years, the introduction of LLMs has redefined the need for rule-based NLP systems, enabling the capture of nuanced context surrounding an abbreviation for disambiguation with near-perfect zero-shot accuracy.
%Modern LLMs can achieve near-perfect zero-shot accuracy by interpreting the nuanced context surrounding an abbreviation. 
Furthermore, sophisticated techniques such as Retrieval Augmented Generation~\cite{rag} (RAG) and dynamic prompting~\cite{few_shot, cot, sandbagging, emotionprompt, rephrase, rereading, expertprompting} have demonstrated that context injection can solve highly specialized disambiguation tasks even without extensive domain-specific fine-tuning.

However, a critical barrier prevents the widespread clinical deployment of these cutting-edge models as production-ready solutions. The highest performing LLMs predominantly operate using cloud-based APIs and proprietary architecture/model weights. Transmitting unredacted, sensitive clinical narratives which contain Protected Health Information (PHI) to these external servers introduces potential security vulnerabilities and frequently violation of regulatory frameworks like Health Insurance Portability and Accountability Act (HIPAA) and General Data Protection Regulation (GDPR). While preemptive de-identification is a common practice~\cite{deindent_0, deindent_1, deindent_2, deident_3}, it is notoriously brittle and aggressive redaction of a clinical note often removes the precise local context the model requires to accurately disambiguate the acronym in the first place.

% To bridge the gap between high-performance natural language understanding and strict data privacy, this paper provides investigation on a novel, privacy preserving pipeline for clinical acronym disambiguation. Rather than relying on large LLMs that uses external APIs, we implement a local-first deployment. By leveraging highly optimized open-weight models executing locally, we guarantee that sensitive clinical information is never leaked. Furthermore, we adapt a simple straightforward zero-shot prompting approach to replicate real-world complexity when providing limited input information to the models.

To bridge the gap between high-performance natural language understanding and strict data privacy, this paper provides investigation on a novel, privacy preserving pipeline for clinical acronym disambiguation. Rather than relying on LLMs that uses external APIs, we implement a local-first deployment. This decentralized approach obviates the requirement for continuous internet connectivity, ensuring uninterrupted functionality in secure clinical environments. By leveraging highly optimized open-weight models executing locally, we enhance the security and confidentiality of sensitive clinical information. Moreover, adopting these small local models resolves the substantial environmental issues and massive energy consumption associated with high-performance server-based systems, such as OpenAI and Google LLMs. Furthermore, we adapt a simple straightforward zero-shot prompting approach to replicate real-world complexity when providing limited input information to the models.

The main contributions of this work are as follows:
\begin{enumerate}\vspace{-1em}
    \item We introduce a novel, fully localized, privacy-preserving LLM-based pipeline specifically designed for acronym disambiguation in sensitive clinical narratives
    \item We demonstrate the efficacy of utilizing the current off-the-shelf small models combined with zero-shot prompting for handling clinical narrative and disambiguation of the complex acronyms
    \item We provide a comprehensive evaluation comparing various small models for when inferenced locally
\end{enumerate}

%% file: related_work.tex
\section*{Related Work}
\vspace{-0.5em}The foundational methodologies for computational sense disambiguation, also known as Word Sense Disambiguation (WSD), were established in the late 1980s and 1990s. WSD refers to the task of selecting the correct meaning of a word occurrence in text from a set of possible meanings provided by a sense inventory. Early experiments by Black~\cite{black} and Hearst~\cite{hearst}, established the groundwork for computational discrimination of English word senses, while Gale et al.~\cite{yarowsky1995}, demonstrated that unsupervised clustering of context vectors could achieve supervised-level performance. As patient safety concerns and medical record digitization grew in the early 2000s, WSD research shifted toward the biomedical domain, motivated by the error analyses of Kopec et al.~\cite{kopec2003human} and Ash~\cite{ash2004some}. Major early milestones included Aronson's~\cite{aronson2001effective} release of MetaMap for mapping text to the UMLS metathesaurus, and Moon et al.'s~\cite{moon2012automatic} comprehensive sense inventory for clinical abbreviations.

Driven by these safety necessities, the late 2000s and early 2010s saw the development of specialized statistical methods for medical shorthand. Following Kuhn’s~\cite{kuhn} warnings on abbreviation risks and Chemali et al.'s~\cite{chemali2015general} findings on general practitioner comprehension gaps, researchers focused on targeted solutions. Key advancements included Joshi et al.'s~\cite{joshi2006comparative} supervised learning approaches, Moon et al.'s~\cite{moon2014sense} standardized sense inventory, and Xu et al.'s~\cite{xu2012combining} hybrid method combining sense-profile vectors with clustering-derived frequency estimates.

Research later expanded into sophisticated contextual modeling and its direct impact on downstream clinical tasks. Chasin et al.~\cite{chasinword} evaluated unsupervised clinical WSD, Li et al.~\cite{li2015acronym} utilized word embeddings to bypass domain-specific knowledge bases, and Wu et al.~\cite{wu2017long} introduced the clinical abbreviation recognition and disambiguation (CARD) framework, outperforming legacy systems like MetaMap. Clinical context was further defined by formally distinguishing expansion from disambiguation~\cite{kim2020improved}, analyzing physician usage frequencies~\cite{hamiel2018frequency}, and identifying unexplained abbreviations and inter-clinic differences in discharge summaries~\cite{schwarz2021structure}. Highlighting the downstream value of this work, Nakayama et al.~\cite{nakayama2019making} showed that normalizing abbreviations significantly improved mortality predictions. Similarly, Wen~\cite{wen2020medal} released the 14.4 million-article MeDAL corpus, demonstrating that pretraining on abbreviation disambiguation accelerates convergence and enhances downstream clinical prediction accuracy.

Most recently, deep learning and transformer architectures have established new benchmarks in acronym resolution. Early hybrid and neural methods~\cite{kashyap2020classe, link2020} were quickly followed by advanced representations, such as Adams et al.'s~\cite{adams2020} Latent Meaning Cells for zero-shot expansion and Skreta et al.'s~\cite{skreta2021automatically} ontology-aware hierarchical CNN framework. Amosa et al.~\cite{amosa2023clinical} noted that while modern NLP models achieve up to 99\% disambiguation accuracy, real-world deployment remains hindered by data scarcity and bias. Addressing these limitations, Jaber~\cite{jaber2022} introduced a unified classifier that fine-tunes clinical BERT models (BioBERT~\cite{biobert}, BlueBERT~\cite{bluebert}, and MS-BERT~\cite{ms-bert}) to robustly disambiguate both common and rare abbreviations without requiring individual models. However, its broader generalizability and clinical viability remain constrained, the evaluation was limited to three base models with highly imbalanced UMN dataset~\cite{UMC} (75 abbreviations, 348 senses), and the study does not explicitly address if privacy-preserving settings was utilized necessary for processing sensitive patient data.
%transformer-based solutions have proven highly effective where Jaber~\cite{jaber2022} proposed a one-fits-all deep learning classifier that fine-tunes pretrained clinical BERT models (such as BioBERT~\cite{biobert}, BlueBERT~\cite{bluebert}, and MS-BERT~\cite{ms-bert}) on the University of Minnesota abbreviation (UMN) dataset~\cite{UMC} achieving higher performance than previous solution showcasing that a single model can robustly disambiguate both common and rare clinical abbreviation senses without needing separate classifiers for each abbreviations. 
Simultaneously, Kugic et al.~\cite{kugic2024disambiguation} began evaluating zero-shot acronym-disambiguation in clinical narratives using Large Language Models (LLMs) with higher parameters such as; GPT‑3.5, GPT‑4, Llama‑2‑7b‑chat and Llama‑2‑70b‑chat on English (CASI), German and Portuguese datasets.

Despite the extensive evolution of acronym disambiguation techniques from early statistical methods to the current dominance of large language models we observed a critical gap in the literature. While recent studies have demonstrated the high accuracy of zero-shot LLMs and complex retrieval-augmented pipelines across both medical and specialized non-clinical domains, these approaches implicitly rely on cloud-based APIs (bigger models) or resource-intensive external server environments~\cite{jaber2022,kugic2024disambiguation}. However, none of the existing works has addressed the data security and privacy-preserving requirements inherent to processing real-world clinical narratives. 
Considering the urgent need to transition these highly effective disambiguation strategies into fully localized frameworks such as executing open-weight models directly on secure infrastructure, our work focuses on investigating the how can we achieve highest performance accuracy while maintaining a privacy-compliant environment.

%% file: experiment_setup.tex
\section*{Experimental Setup}
\vspace{-0.5em}The main focus for the experimental design in our study is to maintain a privacy-preserving environment for the clinical data security when using LLM models for acronym disambiguation. All the experiments and evaluations are strictly performed on-device~(Apple M4-Max GPU using MLX setup) ensuring zero transmission of data to external or cloud-based APIs. The outcomes of these experiments confirms the current viability of local small models when handling acronym detection and contextual expansion accuracy without risking the data leakage. 
%To evaluate the efficacy of the proposed privacy-preserving acronym disambiguation framework, we designed a series of experiments to assess the acronym detection and contextual expansion accuracy. 
This section describes the process pipeline, dataset, models, prompting strategies, evaluation strategies used to test the acronym disambiguation across the clinical domain in a strictly privacy-preserving environment. We approximately benchmark the performance of the off-the-shelf local small models and executed them via Apple Inc's MLX framework~\cite{mlx2023} for GPU inferencing. 
%By executing model inference strictly within a localized environment (Apple M4-Max GPU using MLX or MPS setup), we inherently mitigate the risk of exposing protected health information (PHI) to external APIs. 
Model performance is quantified by initially comparing the accuracy of both the acronym detection and the expansion of the detected acronym and later only the expansion for a given input instance along with the acronym against human annotated ground-truths (GTs) provided by the GLADIS (General and Large Acronym Disambiguation Benchmark)~\cite{gladis}.

\begin{figure}[H]
  \centering
    \includegraphics[width=1.0\textwidth]{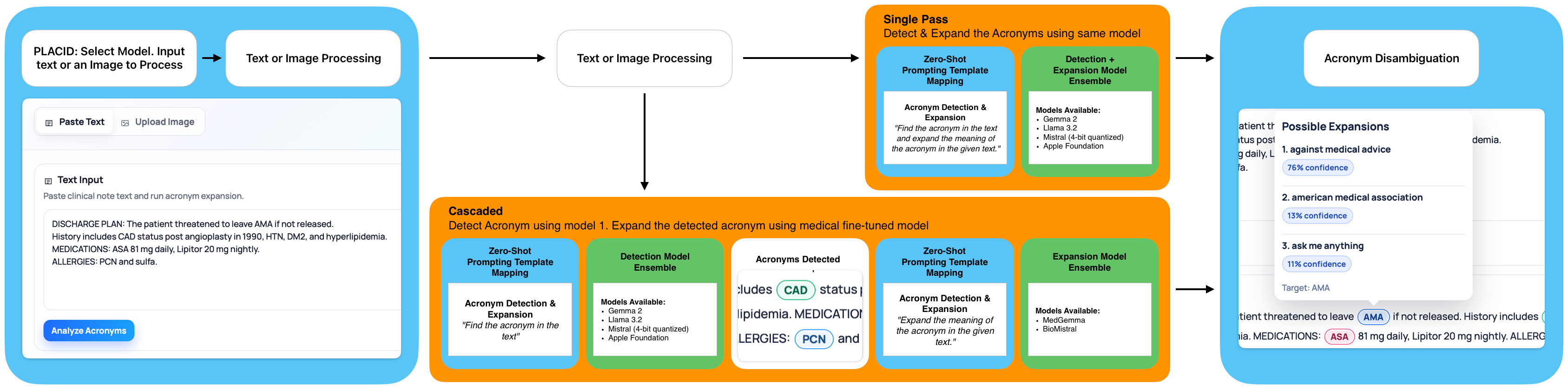}
 \caption{PLACID multi-stage process pipelines for clinical acronym disambiguation using local small models. As illustrated, inference can be deployed in either a single-pass (simultaneous acronym detection and expansion via a single model) or cascaded (general model for detection followed by a medically fine-tuned model for expansion) configuration, depending on local computational constraints.}
 \label{fig:pipeline}
\end{figure}

\subsection*{Process Pipeline}\label{pipeline}
\vspace{-0.5em}We implemented a multi-stage, zero-shot inference pipeline for acronym disambiguation operating on clinical narratives. Our pipeline is designed for an 
%combined working of general purpose foundation models with domain-specific medically trained language models, the 
architecture that yields a clinical expansion of acronyms along with the expansion confidence and rationale of the expansion as illustrated in Figure~\ref{fig:pipeline}. 
%This system is coordinated via prompting template mapping, utilizing two distinct templates i.e., an acronym detection prompt instructing the model to detect the acronyms in the given instance, and an expansion prompt instructing it to generate the contextually relevant expansion of the acronym. 
To balance computational efficiency with output fidelity, the pipeline supports two distinct operational models. The single-pass model allows one model to independently execute both acronym detection and expansion, offering a high computationally efficient end to end approach requiring only one inference call per input, making it ideal for deployments with strict latency constraints. The cascaded specialization mode utilizes a separation of tasks based on model strengths, where general-purpose small model first detects the acronyms, then the detected acronym is routed alongside the original text to the domain-specific model for relevant expansion generation. 

Specifically, the cascaded mode workflow proceeds in four sequential steps, beginning with the processing of a clinical text sample by the prompt template mapping to generate a zero-shot acronym detection prompt. This prompt is sent to the general-domain small model which processes the input instance and determines the acronyms present in the given text. Following this step, the detected acronym(s) are sent to the prompt mapping along with the original input text to construct an expansion prompt, reformulating the task from open-ended identification to focused disambiguation and expansion generation. Finally, this prompt is routed to a domain-specific model. Leveraging its biomedical pretraining, this specialized model generates a clinically grounded expansion for the acronym and a self-assessed confidence score in the range of [0, 1] for the expansions. Both the modes for the current design, output a single definitive expansion paired with its corresponding confidence score. This score explicitly quantifies the model's internal certainty regarding the accuracy of its generated expansion. By reflecting the extent to which the model believes its output is correct, the score serves as an essential calibration signal, helping to identify instances where the system might erroneously assign high confidence to speculative or poor-quality expansion.

\vspace{-1em}\subsection*{Dataset}
The foundation of our experimental setup relies on the biomedical subset as well as some portion of general domain of the GLADIS benchmark. To replicate real-world electronics health records (EHR) environments while utilizing an established, rigorously standardized corpus, we specifically isolated the biomedical test and validation splits.

\vspace{-1em}\subsubsection{Characteristics}\vspace{-0.5em}
The biomedical subset of the GLADIS comprises of $\sim$12k samples (6,295 training, 3,150 validation, and 3,149 test) that were obtained by adapting the MedMentions~\cite{medmentions} corpus of PubMed abstracts~\cite{pile}. These documents are characterized by dense domain-specific jargon and syntactic structures that significantly compound the difficulty of acronym disambiguation. To ensure clinical validity and rigorous standardization, the dateset inherits its long-form entries directly from the Unified Medical Language System (UMLS)~\cite{umls} knowledge base, linking each ambiguous short form to an established biomedical concept. Following strict rule-based filtering, the vast majority of the long forms are classified as clean, providing a high-quality reliable dictionary necessary for robust contextual acronym disambiguation.

\vspace{-1em}\subsubsection{Preparation}\label{preparation}
% The data preparation pipeline was designed to extract the sentences from the tokens format in GLADIS, short-term (acronyms) in the corresponding text along with the annotated expansion ground-truth. 
\vspace{-0.5em}The data preparation in our work was designed to map the clinical text into prompt-ready inputs suitable for generative inference. The parsing of biomedical validation and test data of the GLADIS dataset yielded a total of 6,299 unique text inputs~\footnote{This curated GLADIS biomedical CSV file will be released upon the publication of the paper} with the acronyms \& their corresponding expansion. From this initial pool, a subset of single acronym per input text was separated for the acronym detection accuracy test. 

% To strictly prevent multi-acronym inference during model's acronym detection evaluation, the separation of single acronym subset pool was performed as an additional part of data preparation. 

A simple rule of two or more uppercase characters was used ($\backslash b[A-Z]\{2,\}\backslash b$) that resulted in a total of 2,544 inputs with single acronym per text from the total of 6,299 dataset. Though this method used an approximated rule to determine total number of acronyms in an input text and list the detected acronyms, some input text with equations or non-capitalized acronyms were not classified. Hence, further cleaning of the 2,544 subset was needed to ensure that each input text possesses only one acronym. To perform this process, we used an LLM as an annotator for the nuanced cleaning of the curated subset. 
%detection ability was used to filter the data. 
We used a zero-shot prompting method and asked LLMs (Apple foundation model \& Gemma2:2B) to detect and list all possible acronyms, equations, and alphanumeric characters in the input text. 
%This LLM annotated data was used in cleaning the nuanced acronyms that wasn't classified by the simple rule-based method. 
Lastly, the LLM-based annotated data was verified and cleaned (if needed) by a human-annotator resulting in a final subset of 2,026 single acronym per input text. Each finalized instance (6K data of multiple acronyms per input and 2K data with single acronym per text) was formatted into a standardized input-output pair, linking contextual sentence containing one target acronym and it's corresponding human-annotated ground-truth long-form expansion. 

Additionally, a randomly selected 1K general domain subset, was used for model validation. This subset was specifically used to conduct the primary testing of off-the-shelf local (non-medically tuned) small model's overall ability for acronym detection and expansion. Due to the general lack of information regarding training data used for the training of these small models, a general-domain input detection and expansion test was used to examine the performance gap of the same models when it came to adhering to the clinical inputs. 
%this test was performed to 

% Some additional details regarding the dataset distributions is provided in Appendix~\ref{data_distribution}.
%and their ability to be used without fine-tuning for medical domain applications.
%to study the gap that most common SOTA small models has when it comes to medical soundness capability. 

\vspace{-1em}\subsubsection{Distribution \& Characteristics}~\label{data_distribution}
As outlined in the preparation, our work utilizes only the validation and test subset comprising a total of 6,299 instances dedicated strictly to acronym disambiguation evaluation. 
%For the acronym detection \& expansion accuracy assessment, we parsed the initial 6,299 data pool to isolate texts containing exactly one acronym per input text set, yielding a refined set of 2,026 samples. 
For clarity throughout this paper, we designated the original 6,299 instances for cascaded-mode and the strictly filtered, single-acronym subset of 2,026 instances for single-pass mode.

Applying the acronym extraction rule mentioned in Section~\ref{preparation}, we found 13,881 total acronyms existed within 6,299 instances. Table~\ref{tab:distribution} summarizes the core statistical attributes across the 3 datasets (Single-pass, Cascaded, and General) for in-depth understanding of the input dataset. Quantifying the number of unique acronyms allows us to measure acronym frequency and repetition, which serves as a crucial metric for evaluating model consistency in generation of corresponding expansions. Because a single acronym can map to multiple distinct expansions, this analysis helps determine whether the model genuinely disambiguates the acronym based on the input text context rather than relying on prior parametric memorization in zero-shot prompting.
% Determining the unique acronyms provides us the information on the possible repetition in acronyms and this can be used for the study model consistency. Sometimes, same acronyms can have different expansion and understanding if the model is expanding the acronyms per the context of something it is aware of. 

\vspace{1em}\begin{table}[ht]
\caption{Distribution of acronyms, expansions uniqueness, average tokens, and overshadow ratio (detailed definitions provided in Section~\ref{data_distribution}) for parsed Biomedical and General GLADIS~\cite{gladis} data in our study.}\vspace{-0.5em}
\label{tab:distribution}
\begin{center}
\begin{tabular}{|c|c|c|c|c|c|c|c|}
\hline
\textbf{Dataset} & \textbf{Mode} & \textbf{\begin{tabular}[c]{@{}c@{}}Total\\ Instances\end{tabular}} & \textbf{\begin{tabular}[c]{@{}c@{}}Average\\ Tokens\end{tabular}} & \textbf{\begin{tabular}[c]{@{}c@{}}Unique\\ Acronym\end{tabular}} & \textbf{\begin{tabular}[c]{@{}c@{}}Unique\\ Expansion\end{tabular}} & \textbf{\begin{tabular}[c]{@{}c@{}}Overshadowed\\ Instances\end{tabular}} & \textbf{\begin{tabular}[c]{@{}c@{}}Overshadowed\\ Ratio (\%)\end{tabular}} \\ \hline
\multirow{2}{*}{\textbf{Biomedical}} & Single-pass & 2,026 & 25.26 & 671 & 748 & 126 & 6.22 \\ \cline{2-8} 
 & Cascaded & 6,299 & 30.60 & 1,103 & 1,323 & 613 & 9.73 \\ \hline
\textbf{General} & Single-pass & 1,000 & 56.86 & 56 & 57 & 3 & 0.3 \\ \hline
\end{tabular}
\end{center}
\end{table}

``Overshadowed instance", defined as occurrences where the correct, ground-truth expansion of an acronym is not the most statistically frequent expansion within the corpus overall. For example, if the dataset has 100 sentences containing the acronym "MS" and 80 times it meant \textit{Multiple Sclerosis}, 15 times it meant \textit{Mitral Stenosis}, and 5 times it meant \textit{Morphine Sulfate}. The input sentence uses "MS" to mean as \textit{Mitral Stenosis}, that specific sentence is an overshadowed instance. It forces the LLMs to actually read the context rather than lazily guess the most popular answer (Multiple Sclerosis). 
To calculate the overshadowed ratio a simple equation as shown in Eq.~\ref{overshadowed} is used. This metric is referenced from GLADIS~\cite{gladis} which basically determines the dataset difficulty. 
If the dataset has an overshadowed ratio of 0\% or 5\%, it means models can achieve massive accuracy just by memorizing the most common dictionary definitions without doing any actual language processing. In our study, we opted for an overshadow ratio that added mild complexity to these small models, giving us better understanding on the capabilities of these small models when handling clinical data.

\vspace{-1em}\begin{equation}\label{overshadowed}
    Overshadowed ratio = \frac{Total number of Overshadowed Instances}{Total number of Instances}
\end{equation}

\vspace{-1em}\subsection*{Model Selection}
\vspace{-0.5em}To align with the privacy-preserving constraints, our model selection criteria was restricted to small language models within the 2 to 10-billion parameter range. This scale ensures the models can be executed entirely on local hardware maintaining low-latency inference while avoiding the risk of exposing protected medical information to external APIs. To test the impact of proposed cascaded-mode pipeline we also evaluated general models along with the domain-specific fine-tuned models and examined their collective impact on acronym disambiguation.
%To systematically evaluate how architectural advancements and domain-specific pre-training impact acronym disambiguation, we split our evaluation into 2 distinct model groups, general language/foundational models and biomedical fine-tuned/adapted models.

Initially, we selected 4 state-of-the-art (SOTA) general instruction-following small models, Gemma2:2B~\cite{gemma2}, Llama3.2:3B~\cite{Llama3_2_3b_HuggingFace}, Mistral7B~\cite{mistral}, and Apple Foundation~\cite{apple} model. Gemma2 and Llama3.2 represent highly optimized, parameter-efficient architectures that leverage dense training mixtures to achieve reasoning capabilities traditionally requiring much larger parameter counts. Mistral serves as the upper bound of our evaluated range, providing a robust baseline of context window utilization and instruction following. The Apple Foundation model was integrated in the seamless inference device making it the smallest on-device optimized model.

To quantify the impact of specialized training versus general context reasoning, our extended analysis incorporated additional biomedical fine-tuned models MedGemma1.5:4B~\cite{medgemma}, BioMistral:7B~\cite{biomistral} that are within the same parameter range. These architectures share foundational lineage or parameter scales with the primary group but have undergone extensive fine-tuning, instruction tuning, or alignment specifically on medical datasets such as PubMed abstracts, clinical guidelines, and EHR~\cite{mimic} datasets. BioMistral and MedGemma build upon their respective baseline models by injecting vast quantities of biomedical tokens during fine-tuning phase. By inferencing these domain-specific models with same input instances, we can precisely isolate the performance variance of the general off-the-shelf model when used in clinical narratives. This comparison evaluates whether internalized domain knowledge provided a definitive advantage in resolving acronym disambiguation within limited context.

During our initial model deployments we considered 2 additional medical models HuatuoGPT2:7B~\cite{huatuogpto1medicalcomplexreasoning} and BioGPT~\cite{biogpt} and after careful evaluation we decided to exclude them from our final study. HuatuoGPT2:7B was incompatible with our standardized MLX pipeline due to critical caching and tensor-loading errors that necessitated unacceptable deviations from our localized inference setup. BioGPT (347M parameter base mode) lacked the instruction-following capabilities required to produce structured zero-shot outputs. We will reconsider these models in future studies pending improvements in native framework interoperability, context window expansion, and robust instruction tuning.
%HuatuoGPT2:7B was incompatible with our standardized MLX pipeline, presenting critical caching and tensor-loading errors that required unacceptable deviations from our localized inference setup. BioGPT on the other hand, a 347 million parameter base model, lacked instruction-following capabilities and fundamental ability to produce structured zero-shot outputs. However, with some additional changes on expanded context window, native framework interoperability, and robust instruction-tuning for structured output generation (fine-tuning), these models will be reconsidered in our future studies.

\vspace{-1em}\subsubsection{Model Preparation}
\vspace{-0.5em}All model inference was executed entirely on-device utilizing an Apple Silicon M4-Max architecture. 
%This localized deployment guaranteed a trustworthy privacy-preserving environment for clinical narration. 
To maximize inference efficiency and leveraging the GPU of M4-Max chip, we utilized the MLX framework which is specifically optimized for Apple Silicon natively.

The majority of the selected small models were successfully converted into the MLX format without any quantization. However, deploying the 7-billion Mistral \& BioMistral models presented computational challenges with full precision causing out-of-memory (OOM) errors. To resolve this hardware bottleneck, we converted just the Mistral models to MLX utilizing 4-bit quantization. To ensure that this compression did not degrade the model's contextual reasoning capabilities or introduce hallucination artifacts, we conducted a preliminary validation prior to the deployment. We inferenced the full-precision Mistral against 4-bit quantized counterpart model using a 20 randomly selected input text prompts for acronym detection and expansion operation across 5-iterations. The comparative analysis yielded identical generations across all iterations, demonstrating no degradation in performance accuracy. 

% Additionally, MLX conversion failed for HuatuoGPT-2:7B and BioGPT due to architectural incompatibilities during the conversion process. The conversion pipeline failed due to unsupported model types since the MLX framework currently lack native integration for the baichuan for HuatuoGPT and biogpt architecture. As a result, inference of these model was implemented using backend transformers using "mps" device to endure that these 2 models are using M4-Max GPU while maintaining the default hyperparameters.

% \subsubsection{Local Benchmark} \label{small_models}
% Though our current focus is majorly on the efficiency of the small local models for medical acronym disambiguation, we additionally tested the benchmarks of these models on M4-max silicon processor. This details provided us an in-depth understanding of these models for real-world application deployment when it comes to the common trade-off of the model size to performance accuracy to privacy preserving. 

\vspace{-1em}\subsection*{Prompting Strategies}\label{promting_strategies}
\vspace{-0.5em}For this study, we strictly employed a zero-shot prompting methodology across all the models. While studies indicate other prompt-engineering methods have proven to often improve the LLM generation output~\cite{cot, few_shot, sandbagging, rephrase, rereading,emotionprompt, expertprompting}, it was intentionally excluded from our work. We believe that providing additional details via prompting can artificially inflate a model's performance~\cite{prompt_caution} and this doesn't essentially mimic the real-world scenario of limited input information. Furthermore, by restricting the evaluation strictly to a zero-shot method, we force the models to rely entirely on their trained knowledge base, and their innate ability to perform logical inference based only on the input text provided. This approach rigorously tests the true, out-of-the-box capabilities of these localized models when inputted with high ambiguous, overshadowed acronyms.

The prompt template we used for 2 modes of pipeline are as follows:
\begin{enumerate}\vspace{-1em}
    \item Single-pass mode:
    
    This is a Detect+Expand experiment where only the input text is provided to the model and the model performs an end-to-end operation. For this test, we used the filtered subset of $\sim$2K biomedical and 1K general domain data.
    
    "Task": \textbf{``Find the acronym in the text and expand the meaning of the acronym in the given text."},
    "Text": "input\_text",
    "Rules": ["Output strict JSON on one line"]

    \item Cascaded mode:
    
    This mode follows a two stage process where the list of models first detects the acronyms and then the detected acronym along with the input text is sent as input the stage 2 models. This experiment was performed with an assumption that stage 1 of detection is always accurate and our focus was only testing if the expansions are accurate and contextually relevant. Hence for this experiment, the model received 2 inputs i.e., the acronym along with the text. For this test we used 6K parsed dataset.
    
    "Task": \textbf{``Read the input text carefully, understand the context and expand the meaning of the acronym in the given text."},
    "Text": "input\_text",
    "Rules": ["Output strict JSON on one line"]
\end{enumerate}

This simple template was applied across all the input instances for general \& biomedical domains to analyze the original extent of the models for acronym detection and expansion. Note that for the expansion only experiment, an emphasis on ``understand the context" in the prompt was added since the general models failed in biomedical cases to provide expansions relevant to the input contexts.

% The evaluation metrics selection as well as some additional pre-processing approaches used in our investigation is provided in Appendix~\ref{eval_metrics}
\vspace{-1em}\subsection*{Evaluation Metrics}\label{eval_metrics}
\vspace{-0.5em}Model acronym detection accuracy was evaluated against GLADIS short-term ground truths (GTs) using exact string matching. However, to accommodate morphological variations in small model generated expansions, we measured expansion accuracy using sequence-matching metrics rather than strict string matching. Specifically, we utilized BLEU for n-gram precision (measuring the proportion of generated words that exactly match the ground truth), ROUGE-L for n-gram recall (ensuring the model successfully captures the complete sequence of the expected answer), and METEOR to robustly handle stemming and synonym variations (recognizing when the model generates the correct meaning using slightly different vocabulary). Performance was stratified into high, medium, and low-band accuracy tiers using interpretation thresholds of $\geq$0.7 and $\leq$0.3 across all three metrics. To ensure generative stability, results were averaged across five inference iterations per prompt.

Prior to metric computation, outputs and ground truths underwent two distinct preprocessing pipelines to prevent surface-level formatting conventions from confounding lexical similarity. The "Raw" pipeline applied minimal normalization (Unicode lowercasing and whitespace collapsing) to preserve original punctuation, yielding a conservative similarity baseline. Conversely, the "Clean" pipeline isolated semantic accuracy from arbitrary formatting in complex biomedical nomenclature by additionally removing residual markup and non-alphanumeric characters, and substituting hyphens and underscores with whitespace.

%% file: observations.tex
\section*{Observations}

\vspace{-0.5em}The evaluation of the localized generative models utilizes the dual-mode pipeline~\ref{pipeline} to isolate a model's target acronym detection capabilities within dense clinical narratives from its ability to infer correct long-form expansions. Initially, we assessed the instruction-following capabilities of general, off-the-shelf (non-optimized) small models using a single-pass mode (simultaneous detection and expansion) on a 1k general-domain dataset (Table~\ref{tab:general_perf}). Gemma2:2B demonstrated superior performance in both detection (0.811) and expansion (0.727), followed by the Apple Foundation model (0.785). Iterative evaluations confirmed high response consistency with minimal hallucination. However, high per-row standard deviations (0.369 - 0.484) revealed a bimodal response pattern—inputs were consistently correct or incorrect across iterations with Mistral7B exhibiting the highest polarization and lowest disambiguation accuracy. Expansions mirrored this polarized pattern, marked by high low-band and sparse medium-band proportions, suggesting input-level complexity is the primary performance bottleneck. ROUGE-L consistently outperformed other metrics with a negligible raw-to-clean delta ($\Delta \le 0.002)$, confirming that surface-level formatting does not materially influence outcomes. Finally, the Apple Foundation model's absolute performance was constrained by its built-in safety guardrails, which automatically intercepted and blocked responses, resulting in null outputs. These guardrails are two-layer safety mechanisms that actively scan both user inputs and model outputs to enforce safety policies, automatically intercepting and blocking sensitive, harmful, or inappropriate content before a response is delivered.

\begin{table}[ht]
\caption{Zero-shot local small model's performance accuracy for single-pass mode (detection and expansion) for general-domain 1k input data over 5 iterations. Note, blue text highlights the best scores.}\vspace{-1.5em}
\begin{center}
\label{tab:general_perf}
\begin{tabular}{ccc|ccc|}
\cline{4-6}
\multicolumn{1}{l}{}                       &                                    & \multicolumn{1}{l|}{}                                        & \multicolumn{3}{c|}{\textbf{Expansion Accuracy}}                                              \\ \hline

% \multicolumn{1}{l|}{}                      & \multicolumn{5}{c|}{All Iterations}                                                                                                                                                                         \\ \hline
\multicolumn{1}{|c|}{\textbf{Model}}       & \multicolumn{1}{c|}{\textbf{Text}} & \multicolumn{1}{c|}{\textbf{\begin{tabular}[c]{@{}c@{}}Det.\\ Acc.\end{tabular}}} & \multicolumn{1}{c|}{\textbf{BLEU}} & \multicolumn{1}{c|}{\textbf{METEOR}} & \textbf{ROUGE-L} \\ \hline
\multicolumn{1}{|c|}{\textbf{Foundation}}  & \multicolumn{1}{c|}{Raw}           & \multicolumn{1}{c|}{0.785 ± 0.369}                                                & \multicolumn{1}{c|}{0.654}         & \multicolumn{1}{c|}{0.655}           & 0.680            \\ \hline
\multicolumn{1}{|c|}{\textbf{Foundation}}  & \multicolumn{1}{c|}{Clean}         & \multicolumn{1}{c|}{0.785 ± 0.369}                                                & \multicolumn{1}{c|}{0.654}         & \multicolumn{1}{c|}{0.655}           & 0.680            \\ \hline
\multicolumn{1}{|c|}{\textbf{Gemma2:2b}}   & \multicolumn{1}{c|}{Raw}           & \multicolumn{1}{c|}{\color{blue}{0.811 ± 0.392}}                                                & \multicolumn{1}{c|}{\color{blue}{0.699}}         & \multicolumn{1}{c|}{\color{blue}{0.704}}           & \color{blue}{0.729 }           \\ \hline
\multicolumn{1}{|c|}{\textbf{Gemma2:2b}}   & \multicolumn{1}{c|}{Clean}         & \multicolumn{1}{c|}{\color{blue}{0.811 ± 0.392}}                                                & \multicolumn{1}{c|}{\color{blue}{0.699}}         & \multicolumn{1}{c|}{\color{blue}{0.704}}           & \color{blue}{0.727}            \\ \hline
\multicolumn{1}{|c|}{\textbf{Llama3.2:3b}} & \multicolumn{1}{c|}{Raw}           & \multicolumn{1}{c|}{0.687 ± 0.464}                                                & \multicolumn{1}{c|}{0.612}         & \multicolumn{1}{c|}{0.620}           & 0.643            \\ \hline
\multicolumn{1}{|c|}{\textbf{Llama3.2:3b}} & \multicolumn{1}{c|}{Clean}         & \multicolumn{1}{c|}{0.687 ± 0.464}                                                & \multicolumn{1}{c|}{0.611}         & \multicolumn{1}{c|}{0.620}           & 0.644            \\ \hline
\multicolumn{1}{|c|}{\textbf{Mistral:7b}}  & \multicolumn{1}{c|}{Raw}           & \multicolumn{1}{c|}{0.624 ± 0.484}                                                & \multicolumn{1}{c|}{0.581}         & \multicolumn{1}{c|}{0.586}           & 0.606            \\ \hline
\multicolumn{1}{|c|}{\textbf{Mistral:7b}}  & \multicolumn{1}{c|}{Clean}         & \multicolumn{1}{c|}{0.624 ± 0.484}                                                & \multicolumn{1}{c|}{0.581}         & \multicolumn{1}{c|}{0.587}           & 0.606            \\ \hline
\end{tabular}
\end{center}
\end{table}

\begin{table}[]
\caption{Zero-shot local small model's performance accuracy for single-pass mode (detection and expansion) for biomedical-domain $\sim$2k input data over 5 iterations. Note, blue text highlights the best scores.}\vspace{-1.5em}
\begin{center}
\label{tab:biomedical_perf}
\begin{tabular}{ccc|ccc|}
\cline{4-6}
\multicolumn{1}{l}{}                       &                                    & \multicolumn{1}{l|}{}                                        & \multicolumn{3}{c|}{\textbf{Expansion Accuracy}}                                              \\ \hline
% \multicolumn{1}{l|}{}                      & \multicolumn{5}{c|}{All Iterations}                                                                                                                                                                                   \\ \hline
\multicolumn{1}{|c|}{\textbf{Model}}       & \multicolumn{1}{c|}{\textbf{Text}} & \multicolumn{1}{c|}{\textbf{\begin{tabular}[c]{@{}c@{}}Det.\\ Acc.\end{tabular}}} & \multicolumn{1}{c|}{\textbf{BLEU}} & \multicolumn{1}{c|}{\textbf{METEOR}} & \textbf{ROGUE-L} \\ \hline
\multicolumn{1}{|c|}{\textbf{Foundation}}  & \multicolumn{1}{c|}{Raw}           & \multicolumn{1}{c|}{\color{blue}{0.988 ± 0.072}}                                                & \multicolumn{1}{c|}{0.528}         & \multicolumn{1}{c|}{0.541}           & 0.596            \\ \hline
\multicolumn{1}{|c|}{\textbf{Foundation}}  & \multicolumn{1}{c|}{Clean}         & \multicolumn{1}{c|}{\color{blue}{0.988 ± 0.072}}                                                & \multicolumn{1}{c|}{0.541}         & \multicolumn{1}{c|}{0.564}           & 0.598            \\ \hline
\multicolumn{1}{|c|}{\textbf{Gemma2:2b}}   & \multicolumn{1}{c|}{Raw}           & \multicolumn{1}{c|}{0.930 ± 0.255}                                                & \multicolumn{1}{c|}{0.535}         & \multicolumn{1}{c|}{0.546}           & 0.608            \\ \hline
\multicolumn{1}{|c|}{\textbf{Gemma2:2b}}   & \multicolumn{1}{c|}{Clean}         & \multicolumn{1}{c|}{0.930 ± 0.255}                                                & \multicolumn{1}{c|}{0.550}         & \multicolumn{1}{c|}{0.573}           & 0.608            \\ \hline
\multicolumn{1}{|c|}{\textbf{Llama3.2:3b}} & \multicolumn{1}{c|}{Raw}           & \multicolumn{1}{c|}{0.945 ± 0.229}                                                & \multicolumn{1}{c|}{0.518}         & \multicolumn{1}{c|}{0.535}           & 0.588            \\ \hline
\multicolumn{1}{|c|}{\textbf{Llama3.2:3b}} & \multicolumn{1}{c|}{Clean}         & \multicolumn{1}{c|}{0.945 ± 0.229}                                                & \multicolumn{1}{c|}{0.533}         & \multicolumn{1}{c|}{0.558}           & 0.590            \\ \hline
\multicolumn{1}{|c|}{\textbf{Mistral:7b}}  & \multicolumn{1}{c|}{Raw}           & \multicolumn{1}{c|}{0.892 ± 0.311}                                                & \multicolumn{1}{c|}{0.592}         & \multicolumn{1}{c|}{\color{blue}{0.600}}           & \color{blue}{0.655 }           \\ \hline
\multicolumn{1}{|c|}{\textbf{Mistral:7b}}  & \multicolumn{1}{c|}{Clean}         & \multicolumn{1}{c|}{0.892 ± 0.311}                                                & \multicolumn{1}{c|}{0.605}         & \multicolumn{1}{c|}{\color{blue}{0.622}}           & \color{blue}{0.657}            \\ \hline
\end{tabular}
\end{center}
\end{table}

% We further compared the same observations with the biomedical-domain subset (~2k) as input and the model performance is summarized in Table~\ref{tab:biomedical_perf}. For these domain input instances, all models showed a substantial improvement in detection accuracy with the gains of almost +0.203 to +0.268. Yet the relationship between acronym detection accuracy to relevant expansion of detected acronym was unbalanced. Apple's foundation model achieves the highest detection accuracy (0.987) but does not lead on the expansion accuracy. However, when it came for the expansion accuracy over all the metrics Mistral7B demonstrated superior performance despite having lowest detection accuracy. This dissociation indicates that expansion quality in the biomedical setting is governed more by domain specific language models than by the detection reliability. The persistent Low-band expansion proportions across all models (L: 591–886) despite high detection accuracy demonstrate that successful acronym identification is a necessary factor in our proposed pipeline, but insufficient condition for correct disambiguation. This pronounced performance gap highlights that the difficulty encountered with biomedical inputs is a direct manifestation of the general-domain small model's fundamental lack of specialized medical knowledge which critically impairs their capacity to infer accurate clinical narratives.

Performance on a $\sim$2k biomedical-domain subset (Table~\ref{tab:biomedical_perf}) showed substantial detection improvements (+0.203 to +0.268) across all models, but revealed a dissociation between detection and expansion accuracy. While the Apple Foundation model achieved the highest detection accuracy (0.987), Mistral7B delivered superior expansions despite its lower detection rate. This dissociation indicates that expansion quality in the biomedical setting is governed more by domain specific language models than by the detection reliability. Furthermore, the persistent Low-band expansion proportions across all models (L: 591–886) despite high detection accuracy demonstrate that successful acronym identification is a necessary factor in our proposed pipeline, but insufficient condition for correct disambiguation. This pronounced performance gap underscores that general-purpose models fundamentally lack the specialized medical knowledge required to reliably infer clinical narratives.

\begin{figure}[ht]
  \centering
    \includegraphics[width=1.0\textwidth]{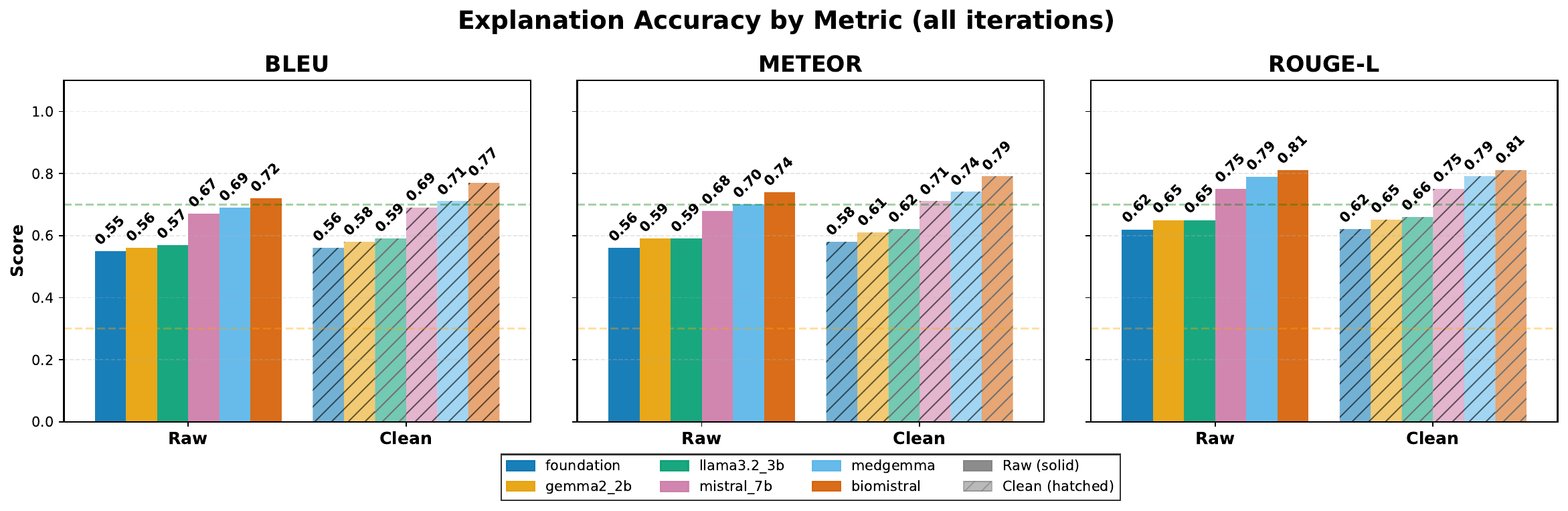}
 \caption{Average expansion accuracy comparison of all models in cascaded-mode over 5 iterations.}
 \label{fig:expansion_only}
\end{figure}

%After establishing the shortcomings of general local models, we know that single-pass mode cannot be used directly for deployment. Hence we moved testing the cascaded-mode and split the model list to general model for acronym detections and expansion of these detected acronyms using medically fine-tuned models.
%the stage 2 (expansion only where the acronym and the input text was provided by us in the prompt) experiments by adding medically fine-tuned models. 
Given the observed shortcomings of off-the-shelf general-purpose local models, relying on a single-pass mode using these models without fine-tuning for clinical deployment proved inadequate. As a result, our evaluation shifted next to testing the cascaded mode. By separating the task, we leveraged the strengths of general models for the initial acronym detection phase, while utilizing medically fine-tuned models to generate accurate, domain-specific expansions for those acronyms. We used the prompting strategy and detection assumption stated in Section~\ref{promting_strategies}. Moreover, Mistral7B has demonstrated superior performance for biomedical domain input, its capabilities are more narrowly specialized rather than broadly generalizable, whereas Gemma2:2B exhibits coherent competence across both domains. Factoring in these observations, we added Gemma \& Mistral medically fine-tuned counterpart models i.e., MedGemma \& BioMistral. The outcomes of the cascaded mode evaluation are summarized in Figure~\ref{fig:expansion_only}. A comparative analysis of expansion accuracy reveals a significant performance difference, with medically fine-tuned models substantially outperforming the general-purpose models. This discrepancy corroborates our underlying hypothesis that the general models leverage broad pattern-recognition capabilities to achieve high detection accuracy, they fundamentally lack the deep, specialized domain knowledge required to generate accurate, contextually grounded medical rationales. Consequently, deploying these models via the cascaded inference mode currently yields the optimal balance of performance. By sequentially leveraging the distinct strengths of these small models within a strict, privacy-preserving local environment, we introduce a novel and highly efficient pipeline for clinical text analysis.

% We also carefully studied the output log of all then model to understand in-depth the model behavior in the output generations. We observed difference in the general model and medically fine-tuned counterparts (such as Gemma vs MedGemma) reasoning approach before the generation of output. To shed more light on the disambiguation mechanism of the domain-specific models, we examine the internal reasoning trace generated during the expansion phase. When presented with an highly ambiguous clinical acronym embedded within its native narrative, the medically fine-tuned model employs a structured evaluation process to weigh competing semantic expansions against the surrounding text. Examples below details the raw inference logs generated by MedGemma during the processing of a sample clinical note containing overdshadowed acronyms resulting to multiple expansions.
We analyzed model output logs to better understand generation behavior, observing distinct reasoning approaches between general models and their medically fine-tuned counterparts (e.g., Gemma vs. MedGemma). To elucidate the disambiguation mechanisms of domain-specific models, we examined their internal reasoning traces during the expansion phase. When processing highly ambiguous acronyms embedded within clinical narratives, fine-tuned models employed a structured evaluation process to weigh competing semantic expansions against the surrounding context. The following examples detail raw MedGemma inference logs, illustrating this process for a sample clinical note containing overshadowed acronyms.

\begin{enumerate}\label{disambiguation_example}\vspace{-1em}
    \item Example 1:
    Input text: "Correlation of PT and famous face and voice recognition was performed."
    
    Response: **Analysis:**
    
    The text mentions "Correlation of PT".PT is a common acronym, but its meaning is context-dependent.
    
    In the context of "famous face and voice recognition", PT could stand for several things:
    1.  **Personal Trainer:** This is a common meaning for PT. However, it doesn't directly relate to face or voice recognition.
    2.  **Physical Therapy:** Similar to Personal Trainer, it's a common meaning but doesn't fit the context.
    3. **Pattern Recognition:** This is the common term used in machine learning but doesn't fit the context.

    \item Example 2:
    Input text: "ED staff cardiac device interrogations are faster , and with similar 30- day outcomes , as compared to SP ."
    
    Response: **Analysis:**
    
    The text mentions "ED staff". ED stands for Emergency Department.
    
    The text also mentions "SP". SP could stand for several things, but without context, it's hard to be certain. However, the context of "cardiac device interrogations" and "30-day outcomes" strongly suggests that SP might refer to a different type of staff or a different department, possibly "SP staff" or "SP department". Given the lack of context, it's difficult to definitively determine the expansion of SP.
\end{enumerate}

\vspace{-1em}As the execution logs illustrate, MedGemma avoids mapping acronyms to its most statistically frequent expansions but instead computed contextual relevance that was not observed in general-models. This transparent, highly calibrated disambiguation process yields both an accurate expansion and an auditable rationale supporting expert clinical reasoning, satisfying critical interpretability requirements for healthcare deployment.

% As illustrated in the execution logs above, MedGemma does not merely map the acronym to its most statistically frequent expansion in general medical literature. Instead, it actively computes contextual relevance which was not observed in general off-the-shelf models. This transparent output demonstrates a highly calibrated disambiguation process that the model provides not only the correct expansion but also an auditable rationale that mimics expert clinical reasoning, satisfying the critical interpretability requirements for healthcare deployment.
%The detailed discussion on the medical model's ability to disambiguate the overshadowed acronyms is provided in Appendix~\ref{AD}

%% file: disucssion.tex
\section*{Discussion \& Future Work}
\vspace{-0.5em}Our experimental results highlight a sharp contrast in the capabilities of current small language models when applied to medical text analysis. Across all evaluations, detection accuracy was consistently high when processing biomedical data. This indicates that general-purpose models possess the necessary linguistic comprehension and instruction-following abilities to accurately isolate and extract the acronyms/abbreviations from the input instances. However, their performance plummeted during the expansion generation phase. When we conducted fair comparison using domain-specific models for the same generative tasks, the expansion accuracy remained robust. This performance gap clearly demonstrates that while general models excel at surface-level pattern recognition and extraction, they fundamentally lack the deep, internalized domain knowledge required to expand accurate, contextually relevant clinical narratives. This deficiency is likely attributable to an under-representation of specialized medical literature and clinical training data. Hence, using a single-pass model approach though computationally efficient, comes at a cost of reduced expansion accuracy if general small models are deployed.  When prompted to self-report confidence scores within their structured JSON outputs, the models frequently exhibited pronounced overconfidence. Due to the black-box nature of these architectures, the internal mechanisms for calculating these metrics remain indeterminable, leading to observed instances where models assigned near-perfect confidence (e.g., 98\%) to fundamentally incorrect expansions

%But when replaced by same model with fine-tuned adapters will solve the expansion accuracy issue.
Some of the observed limitations will be addressed in the future using three primary avenues. First, we plan to address the domain-knowledge deficit by fine-tuning smaller, computationally efficient models on a hybrid dataset. By exposing these models to both highly specialized medical data (such as EHR notes, PubMed abstracts, etc) and general abbreviations (e.g., scientific, organizational, products, etc), we aim to develop a generalized architecture capable of handling the entire acronym detection to expansion pipeline autonomously, reducing the need for complex, multi-model cascading. Second, we will reintroduce and refine an ensemble output methodology for expansion generation. Rather than forcing a single deterministic output, the system will provide users with a prioritized list of expansions by diverse models, each paired with a self-assessed confidence score and an explicit rationale. This approach enables clinical users with differential choices especially in instances where the user is uncertain about a highly ambiguous finding, the model's transparent rationales and confidence scores serve as an interpretive guidance. Finally, we intend to integrate Retrieval-Augmented Generation (RAG) into the pipeline to further bridge the domain-knowledge gap.
%without relying strictly on parametric memory. 
%To rigorously validate the reliability of these metrics, future evaluations will map the ensemble outputs onto calibration plots and consistency heatmaps, ensuring the models are accurately quantifying their own uncertainty rather than confidently hallucinating.

%% file: conclusion.tex
\section*{Conclusion}
\vspace{-0.5em}This study presented a comprehensive evaluation of Large Language Models (LLMs) for acronym disambiguation of clinical narratives, specifically operating within a \textbf{privacy-preserving, local-inference framework} designed to mimic the strict data governance constraints of real-world healthcare deployment. Our empirical results demonstrated a sharp divergence in model capabilities where off-the-shelf general-purpose small models achieved high accuracy (i.e., \textbf{0.988 ± 0.072 for Apple foundation model}) in isolating and detecting acronyms from the input text, their performance degraded significantly during expansion generation, yielding an accuracy of only 0.655 ROGUE-L (highest accuracy for Mistral:7b). Conversely, domain-specific, fine-tuned small models successfully bridged this gap, achieving an \textbf{expansion accuracy of 0.81}. These findings validate our proposed hybrid pipelines of pairing general models for rapid acronym detection with specialized models for disambiguation and expansion generation, which is an effective and secure solution for clinical decision support. Ultimately, this work establishes a baseline for privacy-preserving acronym disambiguation with future optimization, demonstrating that transitioning from cascaded pipeline to a fully fine-tuned, single-pass model will be the critical next step for maximizing both computational efficient and clinical fidelity in on-device deployments.

%% file: amia.bib
@article{berger2026shortcuts,
  title={When shortcuts fall short: The hidden danger of abbreviations in critical care},
  author={Berger, Sebastian and Grzonka, Pascale and Hunziker, Sabina and Frei, Anja I and Sutter, Raoul},
  journal={Journal of Critical Care},
  volume={91},
  pages={155236},
  year={2026},
  publisher={Elsevier}
}

@article{deindent_0,
  title={Learning from experience: privacy and the secondary use of data in health research},
  author={Lowrance, William},
  journal={Journal of health services research \& policy},
  volume={8},
  number={1\_suppl},
  pages={2--7},
  year={2003},
  publisher={SAGE Publications Sage UK: London, England}
}

@article{deindent_1,
  title={Strategies for de-identification and anonymization of electronic health record data for use in multicenter research studies},
  author={Kushida, Clete A and Nichols, Deborah A and Jadrnicek, Rik and Miller, Ric and Walsh, James K and Griffin, Kara},
  journal={Medical care},
  volume={50},
  pages={S82--S101},
  year={2012},
  publisher={LWW}
}

@article{deindent_2,
  title={Methods for the de-identification of electronic health records for genomic research},
  author={El Emam, Khaled},
  journal={Genome medicine},
  volume={3},
  number={4},
  pages={25},
  year={2011},
  publisher={Springer}
}

@article{deident_3,
  title={Combining knowledge-and data-driven methods for de-identification of clinical narratives},
  author={Dehghan, Azad and Kovacevic, Aleksandar and Karystianis, George and Keane, John A and Nenadic, Goran},
  journal={Journal of biomedical informatics},
  volume={58},
  pages={S53--S59},
  year={2015},
  publisher={Elsevier}
}

@article{rag,
  title={Retrieval-augmented generation for knowledge-intensive nlp tasks},
  author={Lewis, Patrick and Perez, Ethan and Piktus, Aleksandra and Petroni, Fabio and Karpukhin, Vladimir and Goyal, Naman and K{\"u}ttler, Heinrich and Lewis, Mike and Yih, Wen-tau and Rockt{\"a}schel, Tim and others},
  journal={Advances in neural information processing systems},
  volume={33},
  pages={9459--9474},
  year={2020}
}

@article{few_shot,
  title={Language models are few-shot learners},
  author={Brown, Tom and Mann, Benjamin and Ryder, Nick and Subbiah, Melanie and Kaplan, Jared D and Dhariwal, Prafulla and Neelakantan, Arvind and Shyam, Pranav and Sastry, Girish and Askell, Amanda and others},
  journal={Advances in neural information processing systems},
  volume={33},
  pages={1877--1901},
  year={2020}
}

@article{cot,
  title={Chain-of-thought prompting elicits reasoning in large language models},
  author={Wei, Jason and Wang, Xuezhi and Schuurmans, Dale and Bosma, Maarten and Xia, Fei and Chi, Ed and Le, Quoc V and Zhou, Denny and others},
  journal={Advances in neural information processing systems},
  volume={35},
  pages={24824--24837},
  year={2022}
}

@inproceedings{sandbagging,
  title={Discovering language model behaviors with model-written evaluations},
  author={Perez, Ethan and Ringer, Sam and Lukosiute, Kamile and Nguyen, Karina and Chen, Edwin and Heiner, Scott and Pettit, Craig and Olsson, Catherine and Kundu, Sandipan and Kadavath, Saurav and others},
  booktitle={Findings of the association for computational linguistics: ACL 2023},
  pages={13387--13434},
  year={2023}
}

@article{emotionprompt,
  title={Emotionprompt: Leveraging psychology for large language models enhancement via emotional stimulus},
  author={Li, Cheng and Wang, Jindong and Zhu, Kaijie and Zhang, Yixuan and Hou, Wenxin and Lian, Jianxun and Xie, Xing},
  journal={arXiv preprint arXiv:2307.11760},
  volume={6},
  year={2023}
}

@inproceedings{rereading,
  title={Re-reading improves reasoning in large language models},
  author={Xu, Xiaohan and Tao, Chongyang and Shen, Tao and Xu, Can and Xu, Hongbo and Long, Guodong and Lou, Jian-Guang and Ma, Shuai},
  booktitle={Proceedings of the 2024 conference on empirical methods in natural language processing},
  pages={15549--15575},
  year={2024}
}

@article{rephrase,
  title={Rephrase and respond: Let large language models ask better questions for themselves},
  author={Deng, Yihe and Zhang, Weitong and Chen, Zixiang and Gu, Quanquan},
  journal={arXiv preprint arXiv:2311.04205},
  year={2023}
}

@article{expertprompting,
  title={Expertprompting: Instructing large language models to be distinguished experts},
  author={Xu, Benfeng and Yang, An and Lin, Junyang and Wang, Quan and Zhou, Chang and Zhang, Yongdong and Mao, Zhendong},
  journal={arXiv preprint arXiv:2305.14688},
  year={2023}
}

@article{kuhn,
  title={Abbreviations and acronyms in healthcare: when shorter isn't sweeter.},
  author={Kuhn, Ivy Fenton},
  journal={Pediatric nursing},
  volume={33},
  number={5},
  year={2007}
}

@inproceedings{gladis,
  title={GLADIS: A general and large acronym disambiguation benchmark},
  author={Chen, Lihu and Varoquaux, Ga{\"e}l and Suchanek, Fabian},
  booktitle={Proceedings of the 17th Conference of the European Chapter of the Association for Computational Linguistics},
  pages={2073--2088},
  year={2023}
}

@article{medmentions,
  title={Medmentions: A large biomedical corpus annotated with umls concepts},
  author={Mohan, Sunil and Li, Donghui},
  journal={arXiv preprint arXiv:1902.09476},
  year={2019}
}

@article{pile,
  title={The pile: An 800gb dataset of diverse text for language modeling},
  author={Gao, Leo and Biderman, Stella and Black, Sid and Golding, Laurence and Hoppe, Travis and Foster, Charles and Phang, Jason and He, Horace and Thite, Anish and Nabeshima, Noa and others},
  journal={arXiv preprint arXiv:2101.00027},
  year={2020}
}

@article{umls,
  title={The unified medical language system (UMLS): integrating biomedical terminology},
  author={Bodenreider, Olivier},
  journal={Nucleic acids research},
  volume={32},
  number={suppl\_1},
  pages={D267--D270},
  year={2004},
  publisher={Oxford University Press}
}

@article{gemma2,
  title={Gemma 2: Improving open language models at a practical size, 2024},
  author={Team, Gemma and Riviere, Morgane and Pathak, Shreya and Sessa, Pier Giuseppe and Hardin, Cassidy and Bhupatiraju, Surya and Hussenot, L{\'e}onard and Mesnard, Thomas and Shahriari, Bobak and Ram{\'e}, Alexandre and others},
  journal={URL https://arxiv. org/abs/2408.00118},
  volume={1},
  number={3},
  year={2024}
}

@misc{Llama3_2_3b_HuggingFace,
  author = {Meta},
  title = {{Llama-3.2-3B} model on Hugging Face},
}

@article{mistral,
  title={Albert q. jiang, alexandre sablayrolles, arthur mensch, chris bamford, devendra singh chaplot, diego de las casas, florian bressand, gianna lengyel, guillaume lample, lucile saulnier, l{\'e}lio renard lavaud, marie-anne lachaux, pierre stock, teven le scao, thibaut lavril, thomas wang, timoth{\'e}e lacroix, william el sayed},
  author={Chaplot, Devendra Singh},
  journal={arXiv preprint arXiv:2310.06825},
  volume={3},
  year={2023}
}

@misc{biomistral,
      title={BioMistral: A Collection of Open-Source Pretrained Large Language Models for Medical Domains}, 
      author={Yanis Labrak and Adrien Bazoge and Emmanuel Morin and Pierre-Antoine Gourraud and Mickael Rouvier and Richard Dufour},
      year={2024},
      eprint={2402.10373},
      archivePrefix={arXiv},
      primaryClass={cs.CL}
}

@article{medgemma,
  title={Medgemma technical report},
  author={Sellergren, Andrew and Kazemzadeh, Sahar and Jaroensri, Tiam and Kiraly, Atilla and Traverse, Madeleine and Kohlberger, Timo and Xu, Shawn and Jamil, Fayaz and Hughes, C{\'\i}an and Lau, Charles and others},
  journal={arXiv preprint arXiv:2507.05201},
  year={2025}
}

@article{apple,
  title={Apple intelligence foundation language models},
  author={Gunter, Tom and Wang, Zirui and Wang, Chong and Pang, Ruoming and Narayanan, Andy and Zhang, Aonan and Zhang, Bowen and Chen, Chen and Chiu, Chung-Cheng and Qiu, David and others},
  journal={arXiv preprint arXiv:2407.21075},
  year={2024}
}

@article{mimic,
  title={MIMIC-III, a freely accessible critical care database},
  author={Johnson, Alistair EW and Pollard, Tom J and Shen, Lu and Lehman, Li-wei H and Feng, Mengling and Ghassemi, Mohammad and Moody, Benjamin and Szolovits, Peter and Anthony Celi, Leo and Mark, Roger G},
  journal={Scientific data},
  volume={3},
  number={1},
  pages={1--9},
  year={2016},
  publisher={Nature Publishing Group}
}

@misc{huatuogpto1medicalcomplexreasoning,
      title={HuatuoGPT-o1, Towards Medical Complex Reasoning with LLMs}, 
      author={Junying Chen and Zhenyang Cai and Ke Ji and Xidong Wang and Wanlong Liu and Rongsheng Wang and Jianye Hou and Benyou Wang},
      year={2024},
      eprint={2412.18925},
      archivePrefix={arXiv},
      primaryClass={cs.CL},
      url={https://arxiv.org/abs/2412.18925}, 
}

@article{biogpt,
    author = {Luo, Renqian and Sun, Liai and Xia, Yingce and Qin, Tao and Zhang, Sheng and Poon, Hoifung and Liu, Tie-Yan},
    title = "{BioGPT: generative pre-trained transformer for biomedical text generation and mining}",
    journal = {Briefings in Bioinformatics},
    volume = {23},
    number = {6},
    year = {2022},
    month = {09},
    issn = {1477-4054},
    doi = {10.1093/bib/bbac409},
    url = {https://doi.org/10.1093/bib/bbac409},
    note = {bbac409},
    eprint = {https://academic.oup.com/bib/article-pdf/23/6/bbac409/47144271/bbac409.pdf},
}

@article{black,
  title={An experiment in computational discrimination of English word senses},
  author={Black, Ezra},
  journal={IBM Journal of research and development},
  volume={32},
  number={2},
  pages={185--194},
  year={1988},
  publisher={IBM}
}

@article{hearst,
  title={Noun homograph disambiguation using local context in large text corpora},
  author={Hearst, Marti},
  journal={Using Corpora},
  pages={185--188},
  year={1991},
  publisher={University of Waterloo, Waterloo Ontario}
}

@inproceedings{yarowsky1995,
  title={Unsupervised word sense disambiguation rivaling supervised methods},
  author={Yarowsky, David},
  booktitle={33rd annual meeting of the association for computational linguistics},
  pages={189--196},
  year={1995}
}

@inproceedings{aronson2001effective,
  title={Effective mapping of biomedical text to the UMLS Metathesaurus: the MetaMap program},
  author={Aronson, Alan R},
  booktitle={Proceedings of the AMIA Symposium},
  pages={17},
  year={2001}
}

@book{moon2012automatic,
  title={Automatic word sense disambiguation of acronyms and abbreviations in clinical texts},
  author={Moon, Sungrim},
  year={2012},
  publisher={University of Minnesota}
}

@article{chemali2015general,
  title={General practitioner understanding of abbreviations used in hospital discharge letters},
  author={Chemali, Mark and Hibbert, Emily J and Sheen, Adrian},
  journal={Medical Journal of Australia},
  volume={203},
  number={3},
  pages={147--147},
  year={2015},
  publisher={Wiley Online Library}
}

@article{kopec2003human,
  title={Human errors in medical practice: systematic classification and reduction with automated information systems},
  author={Kopec, D and Kabir, MH and Reinharth, D and Rothschild, O and Castiglione, JA},
  journal={Journal of medical systems},
  volume={27},
  number={4},
  pages={297--313},
  year={2003},
  publisher={Springer}
}

@article{ash2004some,
  title={Some unintended consequences of information technology in health care: the nature of patient care information system-related errors},
  author={Ash, Joan S and Berg, Marc and Coiera, Enrico},
  journal={Journal of the American Medical Informatics Association},
  volume={11},
  number={2},
  pages={104--112},
  year={2004},
  publisher={BMJ Group BMA House, Tavistock Square, London, WC1H 9JR}
}

@article{chasinword,
  title={Word sense disambiguation in the clinical domain: a comparison of knowledge-rich and knowledge-poor unsupervised methods},
  author={Chasin, Rachel and Rumshisky, Anna and Uzuner, Ozlem and Szolovits, Peter},
  journal={Journal of the American Medical Informatics Association},
  volume={21},
  number={5},
  pages={842--849},
  year={2014},
  publisher={BMJ Publishing Group BMA House, Tavistock Square, London, WC1H 9JR}
}

@inproceedings{li2015acronym,
  title={Acronym disambiguation using word embedding},
  author={Li, Chao and Ji, Lei and Yan, Jun},
  booktitle={Proceedings of the AAAI Conference on Artificial Intelligence},
  volume={29},
  number={1},
  year={2015}
}

@article{wu2017long,
  title={A long journey to short abbreviations: developing an open-source framework for clinical abbreviation recognition and disambiguation (CARD)},
  author={Wu, Yonghui and Denny, Joshua C and Trent Rosenbloom, S and Miller, Randolph A and Giuse, Dario A and Wang, Lulu and Blanquicett, Carmelo and Soysal, Ergin and Xu, Jun and Xu, Hua},
  journal={Journal of the American Medical Informatics Association},
  volume={24},
  number={e1},
  pages={e79--e86},
  year={2017},
  publisher={Oxford University Press}
}

@inproceedings{kim2020improved,
  title={Improved clinical abbreviation expansion via non-sense-based approaches},
  author={Kim, Juyong and Gong, Linyuan and Khim, Justin and Weiss, Jeremy C and Ravikumar, Pradeep},
  booktitle={Machine Learning for Health},
  pages={161--178},
  year={2020},
  organization={PMLR}
}

@article{hamiel2018frequency,
  title={Frequency, comprehension and attitudes of physicians towards abbreviations in the medical record},
  author={Hamiel, Uri and Hecht, Idan and Nemet, Achia and Pe’er, Liron and Man, Vitaly and Hilely, Assaf and Achiron, Asaf},
  journal={Postgraduate Medical Journal},
  volume={94},
  number={1111},
  pages={254--258},
  year={2018},
  publisher={Oxford University Press}
}

@article{nakayama2019making,
  title={Making sense of abbreviations in nursing notes: A case study on mortality prediction},
  author={Nakayama, Jasmine Y and Hertzberg, Vicki and Ho, Joyce C},
  journal={AMIA Summits on Translational Science Proceedings},
  volume={2019},
  pages={275},
  year={2019}
}

@inproceedings{wen2020medal,
  title={MeDAL: Medical abbreviation disambiguation dataset for natural language understanding pretraining},
  author={Wen, Zhi and Lu, Xing Han and Reddy, Siva},
  booktitle={Proceedings of the 3rd clinical natural language processing workshop},
  pages={130--135},
  year={2020}
}

@article{kashyap2020classe,
  title={The CLASSE GATOR (CLinical Acronym SenSE disambiGuATOR): a method for predicting acronym sense from neonatal clinical notes},
  author={Kashyap, Aditya and Burris, Heather and Callison-Burch, Chris and Boland, Mary Regina},
  journal={International journal of medical informatics},
  volume={137},
  pages={104101},
  year={2020},
  publisher={Elsevier}
}

@inproceedings{adams2020,
  title={Zero-shot clinical acronym expansion via latent meaning cells},
  author={Adams, Griffin and Ketenci, Mert and Bhave, Shreyas and Perotte, Adler and Elhadad, No{\'e}mie},
  booktitle={Machine Learning for Health},
  pages={12--40},
  year={2020},
  organization={PMLR}
}

@article{link2020,
  title={Acronym disambiguation in clinical notes from electronic health records},
  author={Link, Nicholas B and Huang, Selena and Cai, Tianrun and He, Zeling and Sun, Jiehuan and Dahal, Kumar and Costa, Lauren and Cho, Kelly and Liao, Katherine and Cai, Tianxi and others},
  journal={medRxiv},
  pages={2020--11},
  year={2020},
  publisher={Cold Spring Harbor Laboratory Press}
}

@article{jaber2022,
  title={Disambiguating clinical abbreviations using a one-fits-all classifier based on deep learning techniques},
  author={Jaber, Areej and Mart{\'\i}nez, Paloma},
  journal={Methods of Information in Medicine},
  volume={61},
  number={S 01},
  pages={e28--e34},
  year={2022},
  publisher={Georg Thieme Verlag KG}
}

@article{UMC,
  title={A sense inventory for clinical abbreviations and acronyms created using clinical notes and medical dictionary resources},
  author={Moon, Sungrim and Pakhomov, Serguei and Liu, Nathan and Ryan, James O and Melton, Genevieve B},
  journal={Journal of the American Medical Informatics Association},
  volume={21},
  number={2},
  pages={299--307},
  year={2014},
  publisher={BMJ Publishing Group BMA House, Tavistock Square, London, WC1H 9JR}
}

@article{biobert,
  title={BioBERT: a pre-trained biomedical language representation model for biomedical text mining},
  author={Lee, Jinhyuk and Yoon, Wonjin and Kim, Sungdong and Kim, Donghyeon and Kim, Sunkyu and So, Chan Ho and Kang, Jaewoo},
  journal={Bioinformatics},
  volume={36},
  number={4},
  pages={1234--1240},
  year={2020},
  publisher={Oxford University Press}
}

@inproceedings{bluebert,
  title={Transfer learning in biomedical natural language processing: an evaluation of BERT and ELMo on ten benchmarking datasets},
  author={Peng, Yifan and Yan, Shankai and Lu, Zhiyong},
  booktitle={Proceedings of the 18th BioNLP workshop and shared task},
  pages={58--65},
  year={2019}
}

@misc{ms-bert,
    title = {{MS-BERT}: A domain-specific language model for processing clinical notes of multiple sclerosis patients},
    author = {{NLP4H} and other contributors},
    howpublished = {\url{https://huggingface.co}},
    note = {{Hugging Face} model repository},
}

@inproceedings{joshi2006comparative,
  title={A comparative study of supervised learning as applied to acronym expansion in clinical reports},
  author={Joshi, Mahesh and Pakhomov, Serguei and Pedersen, Ted and Chute, Christopher G},
  booktitle={AMIA annual symposium proceedings},
  volume={2006},
  pages={399},
  year={2006}
}

@article{moon2014sense,
  title={A sense inventory for clinical abbreviations and acronyms created using clinical notes and medical dictionary resources},
  author={Moon, Sungrim and Pakhomov, Serguei and Liu, Nathan and Ryan, James O and Melton, Genevieve B},
  journal={Journal of the American Medical Informatics Association},
  volume={21},
  number={2},
  pages={299--307},
  year={2014},
  publisher={BMJ Publishing Group BMA House, Tavistock Square, London, WC1H 9JR}
}

@inproceedings{xu2012combining,
  title={Combining corpus-derived sense profiles with estimated frequency information to disambiguate clinical abbreviations},
  author={Xu, Hua and Stetson, Peter D and Friedman, Carol},
  booktitle={AMIA annual symposium proceedings},
  volume={2012},
  pages={1004},
  year={2012}
}

@article{skreta2021automatically,
  title={Automatically disambiguating medical acronyms with ontology-aware deep learning},
  author={Skreta, Marta and Arbabi, Aryan and Wang, Jixuan and Drysdale, Erik and Kelly, Jacob and Singh, Devin and Brudno, Michael},
  journal={Nature communications},
  volume={12},
  number={1},
  pages={5319},
  year={2021},
  publisher={Nature Publishing Group UK London}
}

@article{schwarz2021structure,
  title={Structure, content, unsafe abbreviations, and completeness of discharge summaries: a retrospective analysis in a University Hospital in Austria},
  author={Schwarz, Christine Maria and Hoffmann, Magdalena and Smolle, Christian and Eiber, Michael and Stoiser, Bianca and Pregartner, Gudrun and Kamolz, Lars-Peter and Sendlhofer, Gerald},
  journal={Journal of Evaluation in Clinical Practice},
  volume={27},
  number={6},
  pages={1243--1251},
  year={2021},
  publisher={Wiley Online Library}
}

@article{kugic2024disambiguation,
  title={Disambiguation of acronyms in clinical narratives with large language models},
  author={Kugic, Amila and Schulz, Stefan and Kreuzthaler, Markus},
  journal={Journal of the American Medical Informatics Association},
  volume={31},
  number={9},
  pages={2040--2046},
  year={2024},
  publisher={Oxford University Press}
}

@article{amosa2023clinical,
  title={Clinical errors from acronym use in electronic health record: A review of nlp-based disambiguation techniques},
  author={Amosa, Temitope Ibrahim and Izhar, Lila Iznita Bt and Sebastian, Patrick and Ismail, Idris B and Ibrahim, Oladimeji and Ayinla, Shehu Lukman},
  journal={IEEE Access},
  volume={11},
  pages={59297--59316},
  year={2023},
  publisher={IEEE}
}

@software{mlx2023,
  author = {Awni Hannun and Jagrit Digani and Angelos Katharopoulos and Ronan Collobert},
  title = {{MLX}: Efficient and flexible machine learning on Apple silicon},
  url = {https://github.com/ml-explore/mlx},
  version = {0.0},
  year = {2023},
}

@article{prompt_caution,
  title={Take caution in using LLMs as human surrogates},
  author={Gao, Yuan and Lee, Dokyun and Burtch, Gordon and Fazelpour, Sina},
  journal={Proceedings of the National Academy of Sciences},
  volume={122},
  number={24},
  pages={e2501660122},
  year={2025},
  publisher={National Academy of Sciences}
}
